# Fine-Grained Facial Expression Analysis Using Dimensional Emotion Model

‡Feng Zhou, ‡Shu Kong, Charless C. Fowlkes, Tao Chen, *Baiying Lei, *Member, IEEE*

*Abstract*— Automated facial expression analysis has a variety of applications in human-computer interaction. Traditional methods mainly analyze prototypical facial expressions of no more than eight discrete emotions as a classification task. However, in practice, spontaneous facial expressions in a naturalistic environment can represent not only a wide range of emotions, but also different intensities within an emotion family. In such situations, these methods are not reliable or adequate. In this paper, we propose to train deep convolutional neural networks (CNNs) to analyze facial expressions explainable in a dimensional emotion model. The proposed method accommodates not only a set of basic emotion expressions, but also a full range of other emotions and subtle emotion intensities that we both feel in ourselves and perceive in others in our daily life. Specifically, we first mapped facial expressions into dimensional measures so that we transformed facial expression analysis from a classification problem to a regression one. We then tested our CNN-based methods for facial expression regression and these methods demonstrated promising performance. Moreover, we improved our method by a bilinear pooling which encodes second-order statistics of features. We showed that such bilinear-CNN models significantly outperformed their respective baselines.

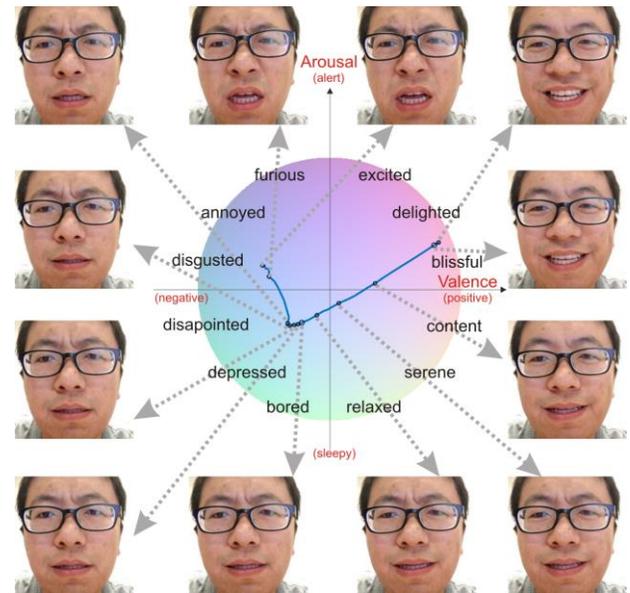

Fig. 1. Our model is trained to embed the face images into the dimensional emotion space which has psychological explainability that enables fine-grained facial expression analysis. In the Valence-Arousal dimensional emotion system, we visualize embeddings output of random frames from a video. We can see the embeddings have smooth transitions along with facial expressions over time.

## I. Introduction

FACIAL expression is one of the main nonverbal cues to convey information between humans, especially emotional information. Thus, it is important to be able to recognize facial expressions, and especially the underlying internal emotional states and intentions in daily life. Automated analysis of facial expressions can introduce the affect dimension into the human-computer interaction process [1] [2] by measuring the level of enjoyment or frustration, which can substantially enhance the user interaction experience by responding to the user in real time accordingly. Nowadays, machine-based automated facial expression recognition has been widely applied in various areas, including human-computer/robot interaction [2], neuromarketing [3], interactive games [1], animations [4], etc.

Recent facial expression analysis research has predominantly focused on recognizing prototypical emotions and the perception of static expressions [5], and has achieved over 95% accuracy on facial expression datasets created under controlled environments (e.g., frontal face, acted facial expressions, homogeneous illumination, and high resolution images) [6]. However, these methods have two main limitations.

First, facial expression datasets, over which these methods are trained, contain only posed expressions, meaning that participants are asked to display different basic emotional expressions (e.g., happiness, sadness, fear, anger, surprise, and disgust) [7]. As a result, these posed facial expressions are not able to include the vast majority of natural facial expressions that humans display and perceive. Thus, these methods are limited in recognizing natural facial expressions by automated agents [8] [9], where facial expressions can be remarkably different from posed ones with regard to intensity, configuration, and duration. In this sense, it is more challenging to analyze natural expressions automatically.

‡Both authors contributed equally to this research.
*Corresponding author.
F. Zhou is with Department of Industrial and Manufacturing Systems Engineering, University of Michigan, Dearborn, 4901 Evergreen Rd., Dearborn, MI 48128, USA (e-mail: fezhou@umich.edu).
S. Kong and C.C. Fowlkes are with Department of Computer Science, University of California, Irvine, 6210, Irvine, CA 92697, USA (email: {skong2, fowlkes}@ics.uci.edu).
T. Chen is with School of Information Science and Engineering, Fudan University, 220 Handan Road, Shanghai, China, 20043. (email:ntuchentao@gmail.com).
B. Lei is with National-Regional Key Technology Engineering Laboratory for Medical Ultrasound, Guangdong Key Laboratory for Biomedical Measurements and Ultrasound Imaging, School of Biomedical Engineering, Health Science Center, Shenzhen University, Shenzhen, China, 518060 (email:leiby@szu.edu.cn)

Second, while it is known that natural facial expressions are dynamically changing under different social interaction circumstances (see Fig. 1), the traditional classification methods are not able to capture the fine-grained variations continuously changing from one emotion to another (e.g., from anger to supersize). Furthermore, there are also subtle variations within one emotion family. One good example is "happy", due to the fact that it is a complicated construct with multiplicity meanings, including positive emotions, life satisfaction, and pleasure [10]. Within the happy family, Fredrickson [11] identified 10 positive emotions, including joy, gratitude, serenity, interest, hope, pride, amusement, inspiration, awe, and love. Humans, in some particular situations, also express multiple or mixed emotions at the same time through facial expressions, (e.g., happily surprised or angrily disgusted, see Fig. 2), especially in the middle of changing displays from one emotion to another [12] [13]. However, the traditional approaches that most studies take, i.e., one facial expression is presented first and then a discrete emotion label has to be assigned, become inadequate in this situation and for natural facial expression recognition as a whole [6] [14]. In this sense, it is of significance to design a new model that is capable of capturing fine-grained differences in dynamic expressions in order to analyze facial expressions.

In this paper, we proposed to analyze fine-grained facial expressions using a dimensional emotion model [15], and transformed the facial expression recognition problem from a classification one into a regression one. Russell [16] proposed a circumplex model that emotional states are collocated in a two-dimensional space, i.e., valence and arousal, termed as core affect [15], and other factors, such as dominance [17], have also been used to explain emotional states. Therefore, unlike traditional approaches, we aim to map facial movements that form facial expressions into the dimensional space. Such a psychological model provides the basis to recognize various types of facial expressions rather than a limited number of basic emotions. Fig. 1 demonstrates the results of facial expression analysis in the dimensional emotion space by our trained model, which is capable of accommodating not only a set of basic emotion expressions, but also a full range of emotions and subtle emotion intensities.

In order to distinguish the subtle variations among different facial expressions, we made use of deep convolutional neural networks (CNNs) to investigate the relationships between facial expressions and emotions represented with dimensional models, i.e., a valence-arousal model and a valence-arousal-dominance model. Compared to traditional machine learning, deep learning has established the new state-of-the-art results in computer vision [18] [12] [19]. Another good reason that we applied a deep CNN for regression is that the input is the image itself without hand-crafted features, which makes it possible to have an end-to-end supervised learning system. In this paper, we proposed to employ bilinear CNNs [20]. Unlike traditional CNNs, bilinear CNNs have two paralleled CNNs and the outputs at their respective locations are fused by taking their matrix outer product. Such second-order features are then pooled to form high-dimensional bilinear feature for classification purposes. They have been proved to be able to distinguish subtle differences among cars, birds, and airplanes [20] [21]. While computationally demanding to obtain bilinear features, it has shown that the low-rank trick can be applied to encode second-order statistics for final classification without explicitly computing the bilinear feature maps [22]. For fine-grained facial expression analysis, we tailored VGG16 [23] and ResNet50 [24] by replacing the softmax loss (used for classification) with a regression loss. We evaluated these two different architectures on a standard FER-2013 dataset [25] with new labels obtained using crowdsourcing [12]. Our results demonstrated that, when pre-trained on large-scale face datasets [26] [27], the deep CNN models achieved very promising results, which were on-par or even better than top teams in Kaggle challenge, which also used FER-2013 as the benchmark; moreover, the models with bilinear pooling significantly outperformed their respective counterparts.

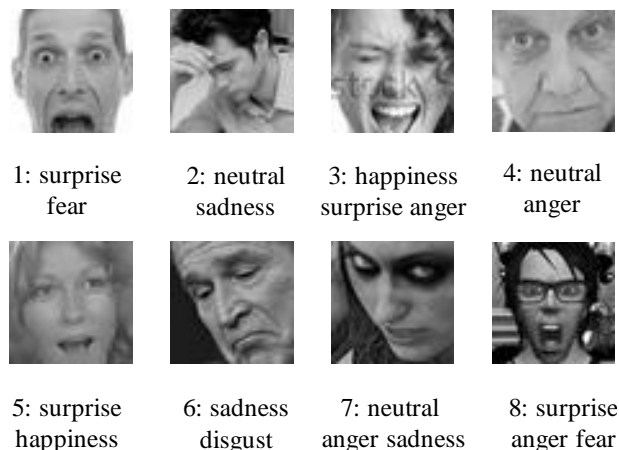

Fig. 2. Multiple emotion labels assigned to facial expressions images from the FER-2013 database [28].

## II. BACKGROUND

### A. Emotion Models

In this research, we focused on two types of emotion models, i.e., discrete and dimensional models. Ekman et al. [29] proposed that emotions were discrete, measurable and physiologically distinct and proposed six basic emotions, including happiness, sadness, anger, disgust, fear, and surprise. From the evolutionary point of view, these emotions have universal facial expressions across cultures [7]. Findings later suggested that contempt was also a basic emotion and its expressions were universally recognized [30]. Although these basic emotions have distinct facial expressions, human expresses many more different forms of facial expressions that are different from those corresponding to basic emotions. Consistent with the discrete emotion concept, Plutchik [31] proposed a wheel of emotions with eight basic emotions, including joy, trust, surprise, anticipation, sadness, fear, anger, and disgust. Unlike

TABLE I
SUMMARY OF PUBLIC FACIAL EXPRESSION DATABASES

| Database | Facial Expressions | Sample Details | Type | Studies |
|---|---|---|---|---|
| CK [46] | neutral, sadness, surprise, happiness, fear, anger, contempt, and disgust | 593 image sequences of 123 subjects | Posed | [41] [42] |
| CK+ [55] | neutral, sadness, surprise, happiness, fear, anger, contempt, and disgust | 500 image sequences of 100 subjects | Posed | [6] [54] |
| JAFFE [47] | neutral, sadness, surprise, happiness, fear, anger, and disgust | 213 static images of 10 subjects | Posed | [6] [42] |
| BU-3DEF [49] | neutral, anger, disgust, fear, happiness, sadness, and surprise | Set I: 2500 static images of 100 subjects; Set II: 606 3D video sequences of 100 subjects; Set III: 2D and 3D videos of 51 subjects | Set I and II: Posed Set III: Spontaneous | Set I and II [48] Set III [56] |
| Bosphorus [50] | happiness, surprise, fear, sadness, anger, disgust | 4652 3D images of 105 subjects | Posed | [50] |
| NVIE [57] | happiness, disgust, fear, surprise, sadness, and anger; arousal and valence on a three-point scale | Spontaneous: Images of 105, 111, and 112 subjects under front, left and right illumination, respectively; Posed: 108 subjects | Spontaneous and posed | [58] |
| AFEW/SFEW [52] | anger, disgust, fear, happy, sadness, surprise, and neutral | SFEW was extracted from AFEW; AFEW: 957 video clips; SFEW: 700 static images | Spontaneous | [51][53] |
| MMI https://mmifacedb.eu/ | happiness, surprise, fear, sadness, anger, disgust | 2900 videos and high-resolution still images of 75 subjects | Posed | [59] |
| SEMAINE [60] | Five affective dimensions and 27 associated categories | 150 participants, for a total of 959 conversations with individual artificial agent characters, lasting approximately 5 minutes each | Spontaneous (between human and artificial agent) | [33] |
| FER-2013 [28] | neutral, sadness, surprise, happiness, fear, anger, contempt, and disgust | 35887 static images of unknown subjects | Spontaneous | [12] |
| FERG [61] | angry, disgust, fear, joy, neutral, sad, and surprise | 55767 annotated face images of six stylized characters | Posed | [61] |

Ekman's basic emotions, these basic emotions could be combined to form another 24 more complex emotions, e.g., love = joy + trust, hope = anticipation + trust, guilt = joy + fear, and contempt = disgust + anger, and so on. Plutchik's model provides further evidence on the variety of facial expressions and facial expressions cannot be attributed to basic emotions only, which are the prevalent methods used in facial expression recognition.

A dimensional model of emotion maps emotions into a two or higher dimensional space. The circumplex model proposed by Russell [16] mapped emotional experiences into a two-dimensional space, i.e., valence and arousal (see Fig. 3). Valence measures how pleasant or unpleasant the emotion is and arousal measures how alert or sleepy the emotion is. For example, happiness is a pleasant emotion (i.e., positive valence) while anger is an unpleasant one (i.e., negative valence). Anger has a high level of arousal while sadness has a low level of arousal. These two dimensions were experimentally supported by Russell using factor analysis of verbal self-report data [32], with additional dimensions equivocal, including dominance, depth of experience, and locus of causation. Valence and arousal were then operationalized as the core affect and capture an important component of emotion [15]. These two dimensions are able to model subtle emotion variations and to differentiate different levels of emotion intensities, for example anger, which ranges from mild irritation to intense fury [33]. Later, Lang et al. [17] added dominance and re-conceptualized it as part of the appraisal process in an emotional episode and they developed standardized affective stimuli (e.g., affective norms for English words [34]) rated against these three dimensions. The dimensional model can accommodate complex mapping relationships between facial expressions and valence, arousal, and (dominance) ratings.

### B. Facial Expression Recognition

Facial expression recognition can be achieved by analyzing facial actions that form the facial expression. The Facial Action Coding Systems (FACS) [35] is probably the earliest system developed to facilitate objective facial expression analysis. The changes in Action Units (AUs) themselves reflect the dynamics of facial expressions. However, the current methods usually only code whether specific AUs are activated or not to indicate if a possible link between basic emotions and facial expressions can be constructed. For example, happiness = activate (AU6 + AU12), sadness = activate (AU1 + AU4 + AU5), and anger = activate (AU4 + AU5 + AU7 + AU23). It is rather cumbersome to use such a manual method. Recently, machine learning methods have been applied to automate facial expression recognition using AUs. For example, Hamm et al. [36] developed an automated FACS by localizing facial landmark with geometric and texture features of video sequences of human faces. Lewinski et al. [37] developed FaceReader to automate facial coding, which reached a FACS index of agreement of 0.70 on two datasets, and showed a reliable indicator of basic emotions. Wegrzyn et al. [38] analyzed the importance of different physical features using FACS and found that the eye and mouth regions, especially when features were grouped, were the most diagnostic ones in successfully recognizing an emo-tion. Such an AU-based method is able to distinguish among different basic emotions. However, it may be inadequate to handle dynamic facial expressions within an emotion family as it abandons the subtle changes of emotions when only a binary AU variable is used.

Another stream of studies extract geometric and appearance-based features to predict facial expressions (and the emotions attributed to these expressions) using machine learning methods. The methods based on geometric features encode location relations of main facial components (e.g., eyes, mouth, and nose) to distinguish different facial expressions. For example, Kotsia and Pitas [39] applied a grid-tracking and deformation system to obtain geometrical displacement of Candide nodes, based on which a support vector machine classifier was able to recognize six basic facial expressions. Xie and Lam [40] proposed a spatially maximum occurrence model to represent facial expressions, and then an elastic shape-texture matching algorithm was applied to predict facial expressions based on the similarity between training examples and testing facial expressions. Geometric features rely on the exact locations of key facial landmarks, which can be difficult to detect with varied illumination, poses, and/or appearances. Methods based on appearances usually extract features from local face regions, such as local binary patterns [41], local directional patterns [42], and histograms of oriented gradients [43], etc., while those extracting features from a global face, such as Eigenfaces [44] and Fisherfaces [45], are mainly used for face recognition. Shan et al. [41] made use of local binary patterns to recognize facial expressions with various machine learning methods for a comparison study, and they showed that support vector machine with boosted local binary patterns obtained best results with around 91.4% accuracy on the Cohn Kanade (CK) database [46]. Jabid et al. [42] used 8 local directional patterns at each pixel and encoded them into an 8 bit binary number to classify facial expressions using template matching and support vector machine. They obtained accuracy around 93.4% on the CK database [46] and 90.1% on the JAFFE database [47]. Dahmane and Meunier [43] applied histograms of oriented gradients to extract appearance features and used support vector machine to recognize 5 discrete emotions, which obtained 70% accuracy.

Recently, deep learning methods have been applied in facial expression recognition due to their great success in computer vision. For example, Li et al. [48] proposed a deep fusion convolutional neural network for multimodal 2D and 3D facial expression recognition, and their method obtained around 86.86% accuracy on BU-3DEF Subset I [49], 81.33% accuracy on BU-3DEF subset II [49], and 80.28% on Bosphorus dataset [50]. Kim et al. [51] trained multiple deep CNNs by adopting several learning strategies and they achieved 61.6% accuracy on the dataset of static facial expression recognition in the wild (SFEW) [52]. Similarly, Yu and Zhang [53] employed an ensemble of multiple deep CNNs and obtained 61.29% on the test set of the SFEW dataset for classifying 7 basic emotions. Meng et al. [54] proposed an identity-aware CNN, which exploited both expression and identity related information for facial expression recognition, and they obtained 95.37% accuracy on the CK+ database [55]. Lopes et al. [6] proposed to combine CNN and specific image pre-processing steps and obtained 96.76% accuracy on the CK+ database [55], 86.74% on the JAFFE database [47], and 91.89% on the BU-3DFE database [49], respectively.

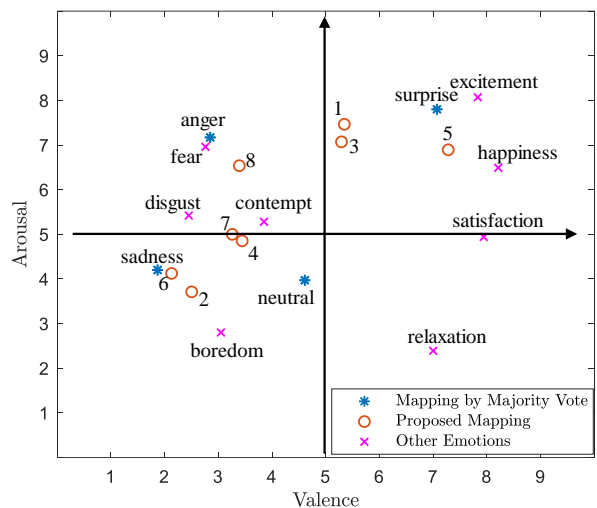

Fig. 3. Dimensional emotion model.

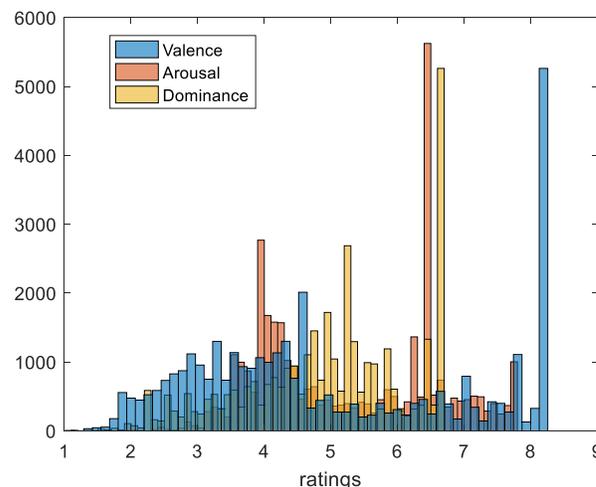

Fig. 4. Distributions of valence, arousal, and dominance ratings of the FER-2013 dataset.

### C. Facial Expression Regression

In order to facilitate natural interactions, dimensional emotion models are used to model many more types of emotions with different levels of variations and intensities. For example, Shin [62] applied a multi-layer perceptron based on features extracted using Gabor wavelet, fuzzy C-means, and dynamic linking model. They evaluated the performance in terms of valence and arousal (representing 44 different emotions) similarity between the predicted results by the model and those by human experts. Zhang et al. [58] investigated the roles of texture and geometry in dimensional facial expression recognition. They found that texture contributed much more than geometry in support vector regression and the best results were obtained using fusion between local binary patterns (i.e., texture fea-tures) and facial animation parameters (geometric features) evaluated on the NVIE database [57] (arousal correlation = 0.498 and valence correlation = 0.690). Banda et al. [33] explored the recognition of continuous dimensional emotion from facial expressions in the context of social human-robot interactions. They proposed a computational model based on recurrent neural networks on a subset of the SEMAINE corpus [60] and labeled the data with valence and arousal with 2 to 8 raters. Their best results had medium correlations between predicted values and ground truth, i.e., arousal correlation = 0.372 and valence correlation = 0.436.

Observing from the studies we reviewed, most of them tested their methods using public facial expression databases with six or seven basic emotions, as summarized in Table I. The majority of the facial expression databases are posed. Compared to the accuracies (usually between 80.28% on Bosphorus dataset and 96.76% on CK+ database) obtained with posed datasets, those obtained from naturalistic datasets were much worse (i.e., 61.6% and 61.29% on SFEW dataset, 71.16% on FER-2013 dataset[1]. Furthermore, not many methods adopted a regression method to distinguish the subtle variations among different emotions. This may be due to the fact that it is extremely costly and subjective to label a facial expression dataset with dimensional values in terms of valence, arousal, and dominance.

### III. Mapping Facial Expressions to Dimensional Emotions

In order to map facial expressions to the dimensional emotion space for the training and testing purpose, we capitalized on the 10 labels obtained by crowd-sourcing for each images from FER-2013 dataset [28]. While more annotators guarantee higher quality of tagging in terms of agreement, 10 annotators are sufficient to reach a high tagging quality that makes machine learning methods able to predict the tags [12]. This tagging agreement also enables us to analyze facial expression through a regression model. Unlike the traditional majority voting method, we made use of each label and converted the facial expressions into valence, arousal, and dominance ratings using affective norms for English words [34] and norms of valence, arousal, and dominance for 13,915 English lemmas [63]. From [34], we obtained valence, arousal, and dominance ratings of seven basic emotions, including happiness, surprise, sadness, anger, disgust, fear, and contempt, and from [63], we obtained the ratings for neutral. They were all rated between 1 and 9, ranging from extremely unpleasant to extremely pleasant for valence, from extremely sleepy to extremely alerted for arousal, and from extremely under control to extremely in control for dominance. The corresponding valence, arousal, and dominance ratings are calculated as follows:

$$r = \frac{HA \cdot r_{ha} + SU \cdot r_{su} + SA \cdot r_{sa} + AN \cdot r_{an} + DI \cdot r_{di} + FE \cdot r_{fe} + CO \cdot r_{co} + NE \cdot r_{ne}}{HA + SU + SA + AN + DI + FE + CO + NE}, \quad (1)$$

[1] www.kaggle.com/c/challenges-in-representation-learning-facial-expression-recognition-challenge/leaderboard

where $r$ is the rating of valence, arousal, or dominance, $HA, SU, SA, AN, DI, FE, CO, NE$ are the numbers of labels of happiness, surprise, sadness, anger, disgust, fear, contempt, and neutral assigned to the facial expression images by crowdsourcing and $r_{ha}, r_{su}, r_{sa}, r_{an}, r_{di}, r_{fe}, r_{co}, r_{ne}$ are the mean ratings provided by [63] and [34]. Take the facial expression images in Fig. 2 as example. These 8 images were assigned to the following labels surprise, sadness, anger, neutral, surprise, sadness, neutral, and anger, provided that a majority voting strategy was used. Using the proposed mapping method, they were mapped from 1 to 8 as shown in Fig. 3. Also other emotions are illustrated in the valence-arousal dimensions based on affective norms for English words. Some emotions are rather close, including anger and fear, 1 and 3, and 4 and 7. Hence, we also incorporated the third dimension, dominance, in this study. Fig. 4 shows the histogram of the ratings of valence, arousal, and dominance. It shows that over 5000 facial expression images with smiles that are labeled with happiness have a high level of agreement among raters from crowdsourcing.

### IV. Dimensional Emotion Regression for Facial Expression Analysis using CNNs

After we mapped facial expressions into dimensional emotions in terms of continuous vectors, we trained deep CNNs to regress towards such continuous embedding vectors. There are many good deep CNN architectures which can be unanimously summarized as a black box, as depicted in Fig. 5 (a). In this paper, we explored two different deep CNN architectures, VGG16 [23] and ResNet50 [24], which are widely studied and exploited in practical applications. The VGG16 network increases its depth using an architecture with 3x3 convolution filters with 16 weight layers, while ResNet50 goes even deeper by introducing skip connections among layers to ease the gradient vanishing/exploding problem, resulting into a 50-layer network. As these architectures were initially designed for object classification which utilizes cross-entropy loss, we modified them accordingly to serve our regression task. Specifically, we removed all the layers after the global pooling layer that aggregated all features spatially; these layers include the last convolution layer and the loss layer. Over the global pooling layer, we stacked a new convolution layer with randomly initialized weights to output final embedding vectors for input images, as depicted by Fig. 5 (b). As for training loss, we turned to a combination of mean-squared error (squared *L2* loss) and mean absolute difference (*L1* loss) as the loss function as shown in Eq. (2) below:

$$Loss = \frac{1}{N} \sum_{i=1}^{N} \left( (Y_i - \hat{Y}_i)^2 + a|(Y_i - \hat{Y}_i)| \right) \quad (2)$$

It is a common practice in literature to use the simplistic *L2* loss for regression. However, we observed that the combination of *L1* and *L2* losses in Eq. (2) consistently helped training. The reason is probably that *L1* loss is more robust to outliers. In our experiments, when setting $a=2$, we achieved stable and consistent improvement compared to using *L2* loss only. Note that we pretrained VGG16 and ResNet50 models on two large-scale face datasets for face recognition task from [26] and [27], respectively.

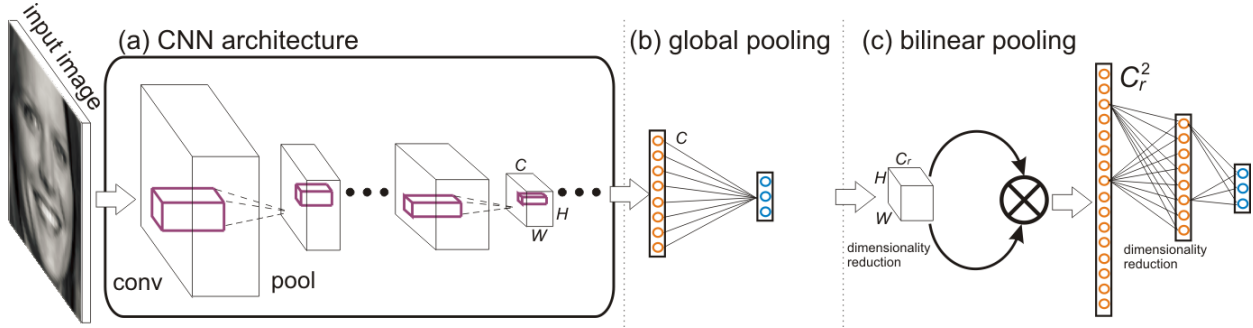

Fig. 5. We trained deep models for dimensional emotion regression to analyze facial expressions. We explored two architectures, VGG16 and ResNet50, as denoted by (a). As a baseline method, we replaced the top layer of the architecture with a global pooling layer that extracted average activations spatially and a new fully-connected layer for regression, as shown in (b). Our improved model exploited bilinear pooling which encoded second-order statistics of feature maps (after a convolution layer for dimensionality reduction) to obtain the holistic feature, which was reduced in dimension through another fully connected layer before the final regression, as shown in (c).

Despite the simple modification as elaborated above, we also improved our model by replacing the global pooling layer with a bilinear pooling one. In literature, bilinear pooling has proved powerful in learning discriminative feature representations that are able to encode second-order statistics [22]. However, the full bilinear pooling feature is very high in dimension that might raise severe memory issues. Nevertheless, Kong and Fowlkes [22] showed that a low-rank constraint can be imposed to reduce the computation for the bilinear feature. While it is straightforward to build a bilinear pooling layer in VGG16 architecture, it would introduce serious computation and memory issues if we build bilinear pooling over the last convolution layer of ResNet50, which would result in a $2048^2$-dimensional feature. Inspired by [22] who advocated the low-rank constraint to reduce dimension and computation, we removed all layers after layer res5c_2 in the last bottleneck residual module in ResNet50, and build the bilinear pooling layer over this layer. This ended up with a final $512^2$-dimensional bilinear feature, just as that in VGG16. Moreover, we found in our experiments that, by inserting one more convolution layer over the bilinear feature with dropout (rate = 0.3) for dimensionality reduction (on both VGG16 and ResNet50), we were able to improve the performance even further.

We evaluated our models on the FER-2013 dataset. Note that some of the images were not rated or removed so the original FER-2013 dataset was reduced from 35887 images to 35714 images, among which 28561 images were used for training, 3579 images for validation, and 3574 images for testing. In the training process, we also augmented the input images with random flip, which has proved to be useful to avoid the overfitting issue [64]. We set the initial learning rate as 0.001 with 0.9 momentum, max epoch as 100 and the learning rate was reduced to 1/2 of the previous one after every 20 epochs. During testing time, we also flipped the image and averaged the prediction to report the performance. Note that, due to the fact that the facial images in the database were gray images with resolution 48 by 48, we resized them to 224 by 224 and concatenated the same three resized gray images along the third dimension to form an RGB image as required by both VGG16 and ResNet50.

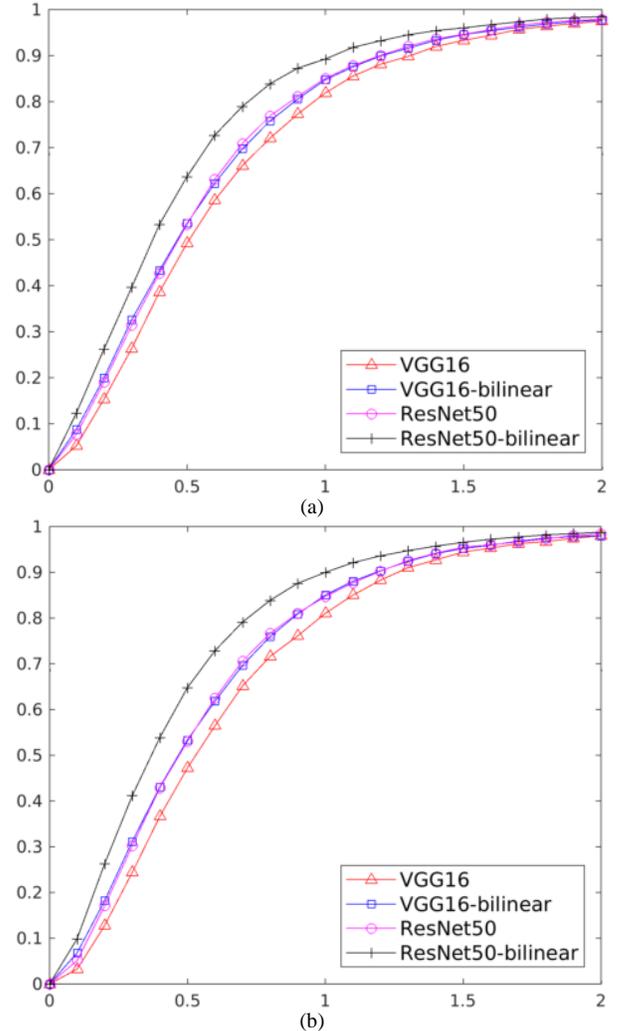

Fig. 6. Prediction accuracy of various models with threshold from 0 to 2. (a) Valence-arousal emotion model and (b) Valence-arousal-dominance emotion model.

## V. RESULTS

We used four measures to evaluate the performance of different models, including root mean squared error (RMSE), Pearson correlation coefficient ($Corr.$), mean absolute error (MAE), and accuracy by setting a threshold (i.e., T = 1) of 50% of the mean standard deviation of valence, arousal, and dominance of affective norms for English words [34] for the eight basic emotions. These measures are defined as follows:

$$RMSE = \sqrt{\frac{1}{N}\sum_{i=1}^{N}(Y_i - \hat{Y}_i)^2}, \quad (3)$$

$$Corr. = \frac{\frac{1}{N}\sum_{i=1}^{N}(Y_i - \bar{Y})(\hat{Y}_i - \bar{\hat{Y}})}{\sqrt{\sum_{i=1}^{N}(Y_i - \bar{Y})^2}\sqrt{\sum_{i=1}^{N}(\hat{Y}_i - \bar{\hat{Y}})^2}}, \quad (4)$$

$$Acc = \frac{\#(Ave\_RMSE < T)}{N}, \quad (5)$$

$$MAE = \frac{1}{N}\sum_{i=1}^{N}|(Y_i - \hat{Y}_i)|, \quad (6)$$

where $N$ is the sample size of testing images of facial expressions, $Y_i$ is the ground truth of valence, arousal, or dominance of the $i$-th testing image, $\hat{Y}_i$ is the predicted value of valence, arousal, or dominance of the $i$-th testing image. Accuracy is defined as the number of accurately predicted images divided by the total number of testing images, and we defined it so that it was accurately predicted if the average RMSE of the valence and arousal or valence, arousal, and dominance was smaller than the threshold. Among these measures, accuracy and correlation are the higher the better, while RMSE and MAE are the lower the better. As shown in Table II and Table III, for the two types of emotion models, both the VGG16-bilinear model and the ResNet50-bilinear model outperformed their corresponding baseline models, which spatially pooled the features into global representation; and the ResNet50 model performed better than the VGG16 model, indicating deeper architecture (ResNet50) learned more powerful features. Specifically, of all the measures in Table II and Table III, ResNet-50-bilinear achieved the best performance. Moreover, by comparing the two types of emotion models, it seemed that the valence-arousal model was slightly better than the valence-arousal-dominance model across different measures. For the valence-arousal model, arousal had smaller errors in terms of RMSE and MAE, and yet its correlation coefficient was smaller compared to valence. For the valence-arousal-dominance model, arousal had the smallest errors in terms of RMSE and MAE, and its correlation coefficient was between valence and dominance. Although dominance had smaller RMSE and MAE compared to valence, its correlation coefficient was the smallest of the three dimensions.

TABLE II
TESTING RESULTS FOR THE VALENCE-AROUSAL EMOTION MODEL WITH TWO TYPES OF ARCHITECTURE (ARCH.) AND TWO POOLING METHODS

| Arch. | pooling | Valence RMSE ↓ | Valence MAE ↓ | Valence Corr. ↑ | Arousal RMSE ↓ | Arousal MAE ↓ | Arousal Corr. ↑ |
|---|---|---|---|---|---|---|---|
| VGG16 | global | 0.92381 | 0.66761 | 0.89536 | 0.69702 | 0.50194 | 0.85973 |
|  | bilinear | 0.87395 | 0.61468 | 0.90713 | 0.65678 | 0.46290 | 0.87674 |
| ResNet50 | global | 0.84067 | 0.59136 | 0.91488 | 0.66772 | 0.47654 | 0.87444 |
|  | bilinear | **0.74111** | **0.50712** | **0.93418** | **0.57488** | **0.39182** | **0.90796** |

TABLE III
TESTING RESULTS FOR THE VALENCE-AROUSAL-DOMINANCE EMOTION MODEL WITH TWO ARCHITECTURES (ARCH.) AND TWO POOLING METHODS

| Arch. | pooling | Valence RMSE ↓ | Valence MAE ↓ | Valence Corr. ↑ | Arousal RMSE ↓ | Arousal MAE ↓ | Arousal Corr. ↑ | Dominance RMSE ↓ | Dominance MAE ↓ | Dominance Corr. ↑ |
|---|---|---|---|---|---|---|---|---|---|---|
| VGG16 | global | 0.95684 | 0.69693 | 0.88706 | 0.72218 | 0.52767 | 0.84867 | 0.71722 | 0.53520 | 0.81796 |
|  | bilinear | 0.90108 | 0.63394 | 0.90054 | 0.67111 | 0.47746 | 0.87071 | 0.68155 | 0.48778 | 0.83787 |
| ResNet50 | global | 0.86386 | 0.61081 | 0.90916 | 0.68851 | 0.49335 | 0.86440 | 0.67653 | 0.48890 | 0.84289 |
|  | bilinear | **0.74420** | **0.50908** | **0.93361** | **0.57892** | **0.39557** | **0.90696** | **0.60678** | **0.42085** | **0.87661** |

TABLE IV
COMPARISONS WITH THE TOP 3 PERFORMANCE TEAMS FROM THE KAGGLE CHALLENGE

| Models | Accuracy | Smallest threshold to achieve 71.16% accuracy |
|---|---|---|
| RBM | 71.16% | - |
| Unsupervised | 69.27% | - |
| Maxim Milakov | 68.82% | - |
| VGG16 Valence-Arousal (T = 1.00) | 81.90% | 0.781 |
| VGG16-bilinear Valence-Arousal (T = 1.00) | 84.75% | 0.718 |
| ResNet-50 Valence-Arousal (T = 1.00) | 85.60% | 0.702 |
| ResNet-50-bilinear Valence-Arousal (T = 1.00) | **89.26%** | **0.582** |
| VGG16 Valence-Arousal-Dominance (T = 1.00) | 81.03% | 0.786 |
| VGG16-bilinear Valence-Arousal-Dominance (T = 1.00) | 85.00% | 0.719 |
| ResNet-50 Valence-Arousal-Dominance (T = 1.00) | 84.72% | 0.707 |
| ResNet-50-bilinear Valence-Arousal-Dominance (T = 1.00) | **89.98%** | **0.576** |

In order to have a direct comparison with classification models reported on Kaggle[2], we set different thresholds of the valence-arousal model and the valence-arousal-dominance model to show the prediction accuracy in Fig. 6. We set the largest threshold as 2 which was the mean standard deviation of valence, arousal, and dominance of affective norms for English words [34] for the eight basic emotions. First, the accuracies increased when the threshold increased and the performance of the valence-arousal model and that of the valence-arousal-dominance was very similar. Second, within each emotion model, the ResNet50-bilinear model achieved the best performance, ResNet50 and VGG16-bilinear had similar performance, and VGG16 had the worst performance. Third, as shown in Table IV, the best classification accuracy was 71.16% achieved by RBM from Kaggle. We listed the accuracy of different models involved in this paper when the threshold was 1 and the smallest threshold that had the same performance achieved by RBM. Consistent with the results in Fig. 6, the best performance was achieved by the ResNet50-bilinear model for both valence-arousal and valence-arousal-dominance model.

---

[2] https://www.kaggle.com/c/challenges-in-representation-learning-facial-expression-recognition-challenge/leaderboard

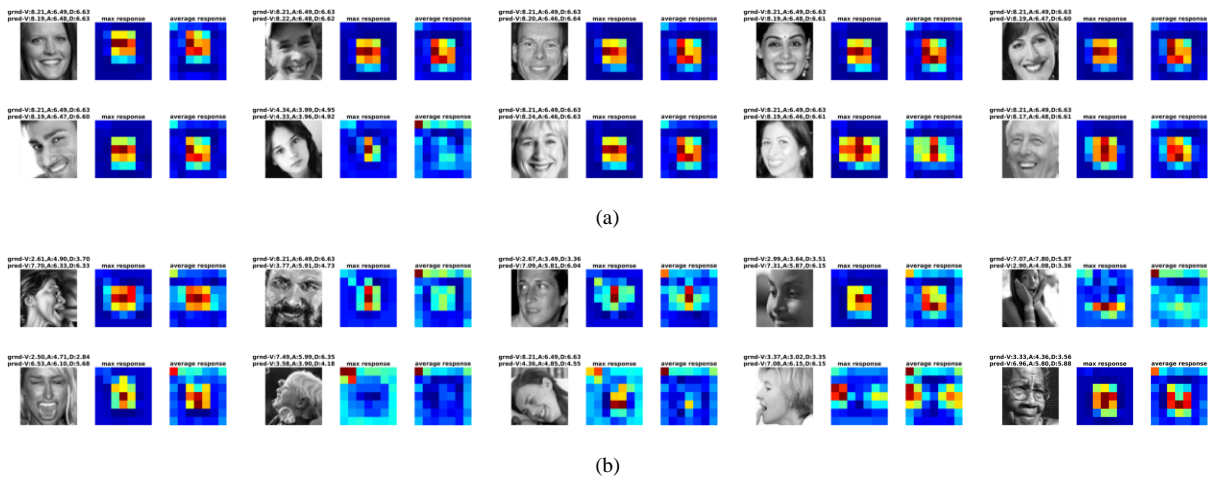

(a)

(b)

Fig. 7. Top 10 best (a) and worst (b) prediction examples on the testing set and their feature maps

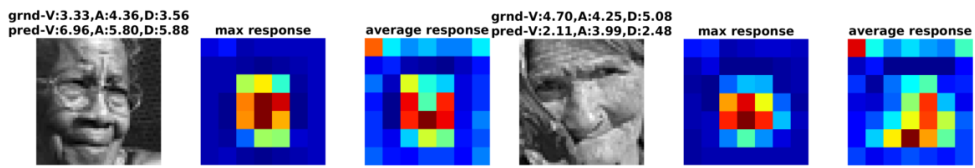

Fig. 8. Bad prediction examples of facial expressions with wrinkles

## VI. Visualization of Feature Map

We summarized the intermediate feature maps of 3D tensor by averaging them along the channel mode into a 2D map. This enabled us to understand what the trained model looked for to analyze facial expressions. Fig. 7 shows the prediction results of eight random examples from the test set with their 2D visualization maps. Generally speaking, these random examples seemed to capture the emotions displayed from their facial expressions and the feature maps were well-centered in the middle, except the second one in the first low, in which the feature map spread out, possibly due to the fact that the face accounted for a smaller area compared to other facial expression images.

In order to further analyze the possible reasons that lead to good or bad prediction results, we further explored the top 10 best prediction results and the top 10 worst prediction results with regard to the valence-arousal-dominance emotion model using the RMSE measure based on the ResNet50-bilinear model as shown in Fig. 8. Note there are two types of feature maps in the table produced by max response and average response. By examining the facial expressions in these two sets, we found that 9 out of the 10 facial expressions had positive valence around 8.21, medium high arousal around 6.49, and medium high dominance around 6.63. This may be due to the fact that there were a large number of facial expressions rated around these values (see Fig. 4 with three peaks of valence, arousal, and dominance). Another reason is that the majority of the face images were frontal and covered nearly over 90% of the image. This was also reflected in the feature maps, which looked compact with regions centered in the middle, possibly due to frontal pose and good registration. On the contrary, the bad examples did not concentrate on specific ranges of valence, arousal, or dominance, which again might be due to the dataset bias (see Fig. 5). Nevertheless, the second and third examples in the second row predicted low valence, arousal, and dominance, which indicated that the model was not entirely biased towards to the most typical examples in the training data. Moreover, the majority of the bad examples had non-frontal or small faces compared with the good examples. This potentially led to the scattered feature maps, showing bad registrations and non-frontal faces.

Another observation, by comparing the top 10 best examples and the top 10 worst examples, was that those in the well predicted group were mainly white, young to mid-aged faces while those in bad predicted group included four black women (i.e., the first, fourth, fifth, and last) with two potentially old (the first and the last). Research [65] has shown such biases exist and they are also potentially caused by the distribution of the training dataset. We further scrutinized the two bad prediction examples with wrinkles (related to age) as shown in Fig. 8. The model potentially picked up the fine-grained textures reflected by the wrinkles in the face images, but failed to predict well due to small numbers of training examples. However, with the current dataset, it was hard to confirm such conjecture and future work is needed to verify this with a new well-annotated large-scale dataset.

## VII. Discussions

**Prototypical facial expressions vs. spontaneous facial expression:** Many researchers [7] [30] agree that there are six or seven basic emotions, including happiness, surprise, fear, anger, contempt, disgust, and sadness. Despite the fact that automated facial expression recognition has achieved accuracies over 95% using dataset with prototypical facial expressions [54] [55] [6], these posed, full-blown, and exaggerated facial expressions often boost the performance of computational models. Another issue with the current research is that these facial expressions are usually labeled as one of the basic emotions as shown in Table I. Such a forced-choice format has been criticized as funneling a range of other facial expressions into these basic emotions [66]. This may explain the different degrees of confusions between basic emotions. For example, fear is frequently confused with surprise (31% [67], 37% [68]), disgust with anger (42% [68], 13% [69]), and sadness with fear (16% [68]) or disgust (15% [69]). Such confusion shows that the boundaries between some basic emotions are not clear [70]. Therefore, we proposed fine-grained facial expression analysis using dimensional emotion models in this research. The boundaries between discrete emotions can be resolved by continuous measurements of valence, arousal, and dominance. Furthermore, the combination of different valence, arousal, and dominance is able to represent many more emotions (that facial expressions can represent) than 6 or 7 basic emotions.

In our daily life, mixed facial expressions are often encountered and prototypical facial configurations disappear, which make facial recognition unreliable [70]. Even within the same basic emotion family, there are different intensity levels, full-blown facial expressions are often better recognized than subtle expressions [71]. For example, Naab and Russell [72] showed 20 spontaneous expressions from Papua New Guinea to 50 subjects. They found that for the 16 facial expressions with only one predicted label, the accuracy ranged from 4.2% to 45.8% with the average 24.2% while for the 4 facial expressions with 2 mixed labels, the accuracies ranged from 6.3% to 66.6% with the average 38.8%. In order to improve the discriminability among spontaneous facial expressions, especially the subtle differences among one emotion family, we applied the bilinear CNN models. Compared to the baseline models (i.e., VGG16 and ResNet50), their corresponding bilinear models significantly performed better in all of the four measures in this research. The main reason the bilinear CNN models are able to capture the fine-grained or subtle distinctions even within an emotion family is that bilinear pooling uses matrix outer product of local features over the whole face image to form a holistic representation [22]. The outer product captures pairwise correlations between the feature channels and can model part-feature interactions [20]. Compared with the best performance from Kaggle in Table IV, our proposed method is able to achieve better performance, when the threshold is above 0.57.

**Limitations and Future Work:** Our study is also limited in several aspects. First, very few databases are labeled with dimensional emotion models. In the current study, we transformed the discrete labels into dimensional models. Thanks to the crowdsourcing labels for the FER-2013 database [12], based on which we converted them into dimensional ratings using affective norms for English words [34]. Nevertheless, the ratings were not directly for the facial expression images, which created extra uncertainties involved in the model. Therefore, databases with reliable ratings on valence, arousal, (and dominance) of different types of faces with better resolutions are needed. Due to the fact that not many studies have explored the fine-grained facial expression analysis and the results from our study are not able to compare with other studies in facial expression recognition directly. The results in Table IV are more of an illustration purpose rather than a strict quantitative comparison.

Second, the performance of the valence-arousal model and the valence-arousal-dominance model is very similar so that we cannot tell if a third dimension, i.e., dominance, is helpful in fine-grained facial expression analysis. However, the prediction results are directly linked to the training set and their dimensional ratings. Therefore, more research is needed to confirm if dominance is needed in fine-grained facial expression analysis.

Third, the results also show that the top 10 best prediction examples are mostly frontal faces, white, and young to middle aged face images while the top 10 worst prediction examples are non-frontal faces, mixed racial, and young to old face images. It is easy to explain the role of pose in fine-grained emotion analysis, and yet the roles of other factors, such as race, gender, and age, are hard to elaborate with the current dataset. Hence, more research is needed in these aspects.

Fourth, although the proposed method is able to tell subtle difference among different facial expressions due to the capabilities of lower-order affective dimensions (i.e., valence, arousal, and dominance), it often needs more information to interpret the emotions associated with such affective dimensions, including semantic knowledge and contextual information [70]. In some other situations, a verbal emotion term (e.g., bored, excited) is better understood than these lower-order affective dimensions.

**Implications for Human-Computer Interaction:** Despite these limitations, our study opens a new perspective to examine automated facial expression recognition. With the transformation from classification to regression using dimensional emotion models, we can create various possibilities for facial expression recognition, which has proved to be not unique to basic emotions, but rather a variety of other emotions (e.g., bored, proud, relaxed, embarrassed), cognitive states (confused, concentrating, distracted), physical states (fatigued, sleepy), and even actions (flirting, looking) [73]. Therefore, it should be cautious to interpret a particular facial expression using just a number of basic emotions, especially for spontaneous facial expressions in our daily life. Rather, the proposed method has the potential to explain a full range of other emotions that individuals both feel in themselves and perceive in others in our daily life. This may greatly improve user experiences of various applications involved in human-computer interaction. For example, not only is the system able to tell if a person feels sad or happy, but it can also recognize if he/she is sleepy when driving, if he/she feels confused when interacting with a new product, and if he/she is bored in a classroom. Furthermore, the bilinear CNN models are able to discriminate subtle changes of facial expressions. This may be used to monitor users' emotional experience continuously when they are interacting with a product. For example, it can be used to monitor a player's facial expression when he/she is playing a game, telling if the user is slightly overwhelmed or greatly overwhelmed, based on which the system can respond accordingly to adjust its difficulty level in order to engage the user better.

## VIII. CONCLUSIONS

In this research, we made use of dimensional emotion models to analyze fine-grained facial expression. Unlike traditional methods that assigned a basic emotion label to a facial expression image, we first mapped facial expression images into a dimensional emotion model, and then proposed a CNN-based deep learning regression models to recognize facial expressions. The results showed that bilinear CNN-based models significantly outperformed their baseline models. Such a method can resolve fundamental issues of traditional facial expression recognition, i.e., complex mapping relationships between emotion terms and facial expressions. The experimental results obtained from the FER-2013 dataset showed the potential and feasibility of the proposed method. In the future, we intend to test this idea with other datasets directly with ratings on facial expressions rather than through affective norms of English words.

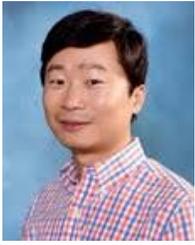

**Feng Zhou** received the bachelor's degree from Ningbo University, Ningbo, China, in 2005, and the M.S. degree from Zhejiang University, Hangzhou, China, in 2007, both in electronic engineering, and the Ph.D. degrees in human factors engineering from Nanyang Technological University, Singapore, in 2011, and in mechanical engineering from the Georgia Institute of Technology, Atlanta, GA, USA, in 2014.

From 2015 to 2017, he was a Research Scientist with Media Science Austin TX. Since Sept. 2017, he has been an Assistant Professor with Department of Industrial and Manufacturing Systems Engineering, University of Michigan, Dearborn. His current research interests include human-computer interaction, engineering design, affective computing, and user research.

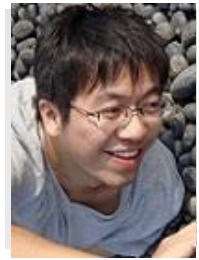

**Shu Kong** received the B.S. degree and M.S. degree both in Computer Science. He is currently a Ph.D. student with the University of California at Irvine. His research interests span computer vision and machine learning, with a focus on solving fine-grained computer vision problems and inter-disciplinary applications to biological and phytological image analysis.

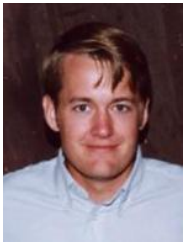

**Charless C. Fowlkes** Charless Fowlkes is an Associate Professor in the Department of Computer Science and the director of the Computational Vision Lab at the University of California, Irvine. Prior to joining UC Irvine, he received his PhD in Computer Science from UC Berkeley in 2005 and a BS with honors from Caltech in 2000.

His research is in computer vision, machine learning and their application to the biological sciences. Dr. Fowlkes is the recipient of the Helmholtz Prize in 2015 for fundamental contributions to computer vision in the area of image segmentation and perceptual grouping, the David Marr Prize in 2009 for his work on contextual models for object recognition, and a National Science Foundation CAREER award. He currently serves on the editorial board of Computer Vision and Image Understanding (CVIU) and IEEE/ACM Transaction on Computational Biology and Bioinformatics.

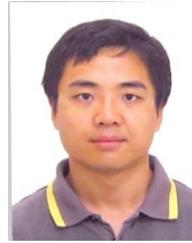

**Tao Chen** rceived the master's degree in information engineering from Zhejiang University in 2008 and the Ph.D. degree in information engineering from Nanyang Technological University in 2013. He is currently a Professor with the School of Information Science and Engineering, Fudan University. His research focuses on computer vision, pattern recognition, and machine learning.

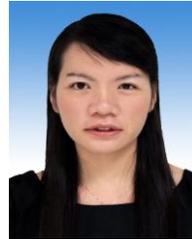

**Baiying Lei** received her M. Eng degree in electronics science and technology from Zhejiang University, China in 2007, and Ph.D. degree from Nanyang Technological University (NTU), Singapore in 2013. She is currently with School of Biomedical Engineering, Shenzhen University, China. Her current research interests include medical image analysis, machine learning, digital watermarking and signal processing.